\documentclass[11pt]{article}
\usepackage{colacl}

\title{Some Properties of Preposition and Subordinate Conjunction
Attachments\thanks{This paper reports on work performed at the MITRE
Corporation under the support of the MITRE Sponsored Research
Program. Useful advice was provided by Lynette Hirschman and David Palmer. The 
experiments made use of Morgan Pecelli's noun/verb group annotations
and some of David Day's programs.
}}

\author{Alexander S. Yeh \and Marc B. Vilain\\ 
        MITRE Corporation\\
        202 Burlington Road\\
        Bedford, MA 01730\\
        USA \\
	\{asy,mbv\}@mitre.org\\
        phone\# +1-781-271-2658
}

\begin{document}
\maketitle
\begin{abstract}
{\begin{picture}(0,0)
\put(0,215){In the 17th International Conference on Computational Linguistics and}
\put(34,200){36th Annual Meeting of the Association for Computational Linguistics}
\put(34,185){(COLING-ACL'98), pages 1436-1442, Montr\'{e}al, Canada. \copyright \/1998 Universit\'{e} de Montr\'{e}al.}
\put(0,-511){cmp-lg/9808007}
\end{picture}}
Determining the attachments of prepositions and subordinate
conjunctions is a key problem in parsing natural language.  This paper
presents a trainable approach to making these attachments through
transformation sequences and error-driven learning.  Our approach is
broad coverage, and accounts for roughly three times the attachment
cases that have previously been handled by corpus-based techniques.
In addition, our approach is based on a simplified model of syntax
that is more consistent with the practice in current state-of-the-art
language processing systems.  This paper sketches syntactic and
algorithmic details, and presents experimental results on data sets
derived from the Penn Treebank.  We obtain an attachment accuracy of
75.4\% for the general case, the first such corpus-based result to be
reported.  For the restricted cases previously studied with
corpus-based methods, our approach yields an accuracy comparable to
current work (83.1\%).
\end{abstract}

\section{Introduction}

Determining the attachments of prepositions and subordinate
conjunctions is an important problem in parsing natural language.  It
is also an old problem that continues to elude a complete solution.  A
classic example of the problem is the sentence {\em ``I saw a man
with a telescope''}, where who had the telescope is ambiguous.

Recently, the preposition attachment problem has been addressed using
corpus-based methods \cite{HandR93,Ratna94,BandR94,CandB95,Merlo97}.
The present paper follows in the path set by these authors, but
extends their work in significant ways.  We made these extensions to
solve this problem in a way that can be directly applied in running
systems in such application areas as information extraction or
conversational interfaces.

In particular, we have sought to produce an attachment decision
procedure with far broader coverage than in earlier approaches.  Most
research to date has focussed on a subset of the attachment problem
that only covers 25\% of the problem instances in our training data,
the so-called binary VNP subset.  Even the broader V[NP]$^*$ subset
addressed by \cite{Merlo97} only accounts for 33\% of the problem
instances.  In contrast, our approach attempts to form attachments for
as much as 89\% of the problem instances (modulo some cases that are
either pathological or accounted for by other means).

Work to date has also been concerned primarily with reproducing the
structure of Treebank annotations.  In other words, the underlying
syntactic paradigm has been the traditional notion of full sentential
parsing.  This approach differs from the parsing models currently
being explored by both theorists and practitioners, which include
semi-parsing strategies and finite-state approximations to
context-free grammars.

Our approach to syntax uses a cascade of rule sequence processors,
each of which can be thought of as approximating some aspect of the
underlying grammar by finite-state transduction.  We have thus had to
extend previous work at the conceptual level as well, by recasting the
preposition attachment problem in terms of the vocabulary of
finite-state approximations (noun groups, etc.), rather than the
traditional syntactic categories (noun phrases, etc.).

Much of the present paper is thus concerned with describing our
extensions to the preposition attachment problem.  We present the
problem scope of interest to us, as well as the data annotations
required to support our investigation.  We also present a decision
procedure for attaching prepositions and subordinate conjunctions.
The procedure is trained through error-driven transformation learning
\cite{BrillPhD}, and we present a number of training experiments and
report on the performance of the trained procedure.  In brief, on the
restricted VNP problem, our procedure achieves nearly the same level
of test-set performance (83.1\%) as current state-of-the-art systems
(84.5\% \cite{CandB95}).  On the unrestricted data set, our procedure
achieves an attachment accuracy of 75.4\%.

\section{Syntactic Considerations}

Our outlook on the attachment problem is influenced by our approach to
syntax, which simplifies the traditional parsing problem in several
ways.  As with many approaches to processing unrestricted text, we do
not attempt as a primary goal to derive spanning sentential parses.
Instead, we approximate spanning parses through successive stages of
partial parsing.  For the purpose of the present paper, we need to
mostly be concerned with the level of analysis of core noun phrases
and verb phrases.  By core phrases, we mean the kind of non-recursive
simplifications of the NP and VP that in the literature go by
names such as noun/verb groups \cite{Appelt93} or chunks,
and base NPs \cite{RandM95}.

The common thread between these approaches and ours is to approximate
full noun phrases or verb phrases by only parsing their non-recursive
core, and thus not attaching modifiers or arguments.  For English noun
phrases, this amounts to roughly the span between the determiner and
the head noun; for English verb phrases, the span runs roughly from
the auxiliary to the head verb.  We call such simplified syntactic
categories {\em groups}, and consider in particular noun, verb, adverb
and adjective groups.

For noun groups in particular, the definition we have adopted also
includes a limited number of constructs that encompass some
depth-bounded recursion.  For example, we also include in the scope of
the noun group such complex determiners as partitives (``five of the
suspects'') and possessives (``John's book'').  These constructs fall
under the scope of our noun group model because they are easy to parse
with simple finite-state cascades, and because they more intuitively
match the notion of a core phrase than do their individual components.
Our model of noun groups also includes an extension of the so-called
named entities familiar to the information extraction community
\cite{Muc6}.  These consist of names of persons and organizations,
location names, titles, dates, times, and various numeric expressions
(such as money terms).  Note in particular that titles and
organization names often include embedded prepositional phrases (e.g.,
``Chief of Staff'').  For such cases, as well as for partitives, we
consider these embedded prepositional phrases to be within the noun
group's scope, and as such are excluded from consideration as
attachment problems. Also excluded are the auxiliary {\em to\/}'s in
verb groups for infinitives.

Once again, distinguishing syntax groups from traditional syntactic
phrases (such as NPs) is of interest because it singles out what is
usually thought of as easy to parse, and allows that piece of the
parsing problem to be addressed by such comparatively simple means as
finite-state machines or transformation sequences.  What is then left
of the parsing problem is the difficult stuff: namely the attachment
of prepositional phrases, relative clauses, and other constructs that
serve in modificational, adjunctive, or argument-passing roles.  This
part of the problem is harder both because of the ambiguous attachment
location, and because the right combination of
knowledge required to reduce this ambiguity is elusive.

\section{The Attachment Problem}\label{s:attachment-problem}

Given these syntactic preliminaries, we can now define attachment
problems in terms of syntax groups.  In addition to noun, verb,
adjective and adverb groups, we also have {\em I-groups}. An I-group
is a preposition (including multiple word prepositions) or subordinate
conjunction (including {\em wh}-words and ``that'').  Once again
prepositions that are embedded in such constructs as titles and names
are not considered I-groups for our purposes. Each I-group in a
sentence is viewed as attaching to one other group within that
sentence.\footnote{Sentential level attachments are deemed to be to
the main verb in the sentence attached to.} For example, the sentence
{\em ``I had sent a cup to her.''}  is viewed as
\begin{center}
[I]$_{ng}$ [had sent]$_{vg,\triangleleft}$ [a cup]$_{ng}$ [to]$_{Ig,\triangleright}$ [her]$_{ng}$.
\end{center}
where $\triangleright$ indicates the attaching I-group and $\triangleleft$ indicates the group
attached to.

Generally, coordinations of groups (e.g., {\em dogs and cats}) are
left as separate groups. However, prenominal coordination (e.g. {\em
dog and cat food}) is deemed as one large noun group.

{\em Attachments not to try:} Our system is designed to attach each
I-group in a sentence to one other group in the sentence on that
I-group's left.  In our sample data, about 11\% of the I-groups have
no left ambiguity (either no group on the left to attach to or only 1
group).  A few (less than 0.5\%) of the I-groups have no group to its
right.  All of these I-groups count as attachments not handled by our
system and our system does not attempt to resolve them.

{\em Attachments to try:} The rest of the I-groups each have at least
2 groups on their left and 1 group on their right from the I-group's
sentence, and these are the I-groups that our system tries to handle
(89\% of all the problems in the data).

\section{Properties of Attachments to Try}
In order to understand how our technique handles the attachments that
follow this pattern, it is helpful to consider the properties of this
class of attachments.  What we detail here is a specific analysis of
our test data (called 7x9x).  Our training sample is similar.

In 7x9x, 2.4\% of the attachments turn out to be of a form that
guarantees our system will fail to resolve them.  83\% of these
unresolvable ``attachments'' are about evenly divided between right
attachments and left attachments to a coordination of groups (which in
our framework is split into 2 or more groups). A right attachment
example is that ``at'' attaches to ``lost'' in \mbox{``that at home,
they lost a key.''} A coordination attachment example is ``with''
attaching to the coordination \mbox{``cats and dogs''} in \mbox{``cats
and dogs with tags''}.  The other 17\% were either lexemes erroneously
tagged as prepositions/subordinate conjunctions or past participles,
or were {\em wh}-words that are actually part of a question (and not
acting as a subordinate conjunction).

In 7x9x, 67.7\% of attachments are to the adjacent group on the
I-group's immediate left. Our system uses as a starting point
the guess that all attachments are to the adjacent group.

The second most likely attachment point is the nearest verb group to
the I-group's left. A surprising 90.3\% of the attachments are to
either this verb group or to the adjacent group.\footnote{This
attachment preference also appears in the large data set used in
\cite{Merlo97}.} In our experiments, limiting the choice of possible
attachment points to these two tended to improve the results and also
increased the training speed, the latter often by a factor of 3 to 4.

Neither of these percentages include attachments to coordinations of
groups on the left, which are unhandleable. Including these
attachments would add $\sim$1\% to each figure.

The attachments can be divided into six categories, based on the
contents of the I-group being attached and the types of groups
surrounding that I-group. The categories are:
\begin{description}
\item [{\bf vnpn}] The I-group contains a preposition. Next to the
preposition on both the left and the right are noun groups. Next to
the left noun group is a verb group.  A member of this category is the
[to]$_{Ig}$ in the sentence ``[I]$_{ng}$ [had sent]$_{vg}$ [a
cup]$_{ng}$ [to]$_{Ig}$ [her]$_{ng}$.''

\item [{\bf vnp\={n}}] Like {\bf vnpn}, but next to the preposition on
the right is not a noun group.

\item [{\bf \={v}npn}] Like {\bf vnpn}, but the left neighbor of the
left noun group is not a verb group.

\item [{\bf \={v}np\={n}}] Another variation on {\bf vnpn}.

\item [{\bf x\={n}px}] The I-group contains a preposition. But its
left neighbor is not a noun group. 
The {\bf x}'s stand for groups that need to exist, but can be of any
type.
 

\item [{\bf xxsx}] The I-group has a subordinate conjunction
(e.g. {\em which}) instead of a preposition.\footnote{A
word is deemed a preposition if it is among the 66 prepositions listed
in Section~\ref{ss:prelim-expr}'s $lt$ data set.  Unlisted words are
deemed subordinate conjunctions.}
\end{description}

Table~\ref{t:vnpn-category} shows how likely the attachments in 7x9x
that belong to each category are 
\begin{itemize}
 \item to attach to the left adjacent group ({\em A})
 \item to attach to either the left adjacent group or the nearest verb
 group on the left ({\em V-A})
 \item to have an attachment that our system actually cannot correctly
 handle ({\em Err}). 
\end{itemize}
The table also gives the percentage of the attachments in 7x9x
that belong in each category ({\em Prev\/}alence). The {\em A} and
{\em V-A} columns do not include attachments to
coordinations of groups.
\begin{table}[htb]
\begin{tabular}{|c||r|r|r||r|}\hline
Category & A & V-A & Err & Prev \\ \hline \hline
{\bf vnpn} & 55.6\% & 97.3\% & 0.8\% & 22.8\% \\ \hline
{\bf vnp\={n}} & 44.4\% & 92.6\% & 0.0\% & 2.4\% \\ \hline
{\bf \={v}npn} & 61.4\% & 85.1\% & 2.5\% & 30.7\%  \\ \hline
{\bf \={v}np\={n}} & 37.7\% & 83.0\% & 3.8\% & 2.4\% \\ \hline
{\bf x\={n}px} & 85.6\% & 93.6\% & 3.3\% & 28.3\%  \\ \hline
{\bf xxsx} & 74.3\% & 84.2\% & 3.3\% & 13.4\%  \\ \hline\hline
Overall & 67.7\% & 90.3\% & 2.4\% & 100\% \\
\hline
\end{tabular}
\caption{Category properties in 7x9x}\label{t:vnpn-category}
\end{table}

Much of the corpus-based work on attaching prepositions
\cite{Ratna94,BandR94,CandB95} has dealt with the subset of category
{\bf vnpn} problems where the preposition actually attaches to either
the nearest verb or noun group on the left. Some earlier work
\cite{HandR93} also handled the subset of {\bf vnp\={n}} category
problems where the attachment is either to the nearest verb or noun
group on the left.

Some later work \cite{Merlo97} dealt with handling from 1 to 3
prepositional phrases in a sentence. The work dealt with prepositions
in ``group'' sequences of VNP, VNPNP and VNPNPNP, where the
prepositions attach to one of the mentioned noun or verb groups (as
opposed to an earlier group on the left). So this work handles
attachments that can be found in the {\bf vnpn}, {\bf vnp\={n}}, {\bf
\={v}npn} and {\bf \={v}np\={n}} categories. Still, this work handles
less than an estimated 33\% of our sample text's
attachments.\footnote{\cite{Merlo97} searches the Penn Treebank for
data samples that they can handle. They find phrases where 78\% of the
items to attach belong to either the {\bf vnpn} or {\bf vnp\={n}}
categories. So in Penn Treebank, they handle 1.28 times more
attachments than the other work mentioned in this paper. This other
work handles less than 25\% of the attachments in our sample data.}

\section{Processing Model}
Our attachment system is an extension of the rule-based system for
VNPN binary prepositional phrase attachment described in
\cite{BandR94}. The system uses transformation-based error-driven
learning to automatically learn rules from training examples.

One first runs the system on a training set, which starts by guessing
that each I-group attaches to its left adjacent group. This training
run moves in iterations, with each iteration producing the next rule
that repairs the most remaining attachment errors in the training
set. The training run ends when the next rule found repairs less than
a threshold number of errors.

The rules are then run in the same order on the test set (which also
starts at an all adjacent attachment state) to see how well they do.

The system makes its decisions based on the head (main) word of each
of the groups examined. Like the original system, our system can look
at the head-word itself and also all the semantic classes the
head-word can belong to. The classes come from Wordnet \cite{Wordnet}
and consist of about 25 noun classes (e.g., person, process) and 15
verb classes (e.g., change, communication, status). As an extension,
our system also looks at the word's part-of-speech, possible stem(s)
and possible subcategorization/complement categories. The latter
consist of over 100 categories for nouns, adjectives and verbs (mainly
the latter) from Comlex \cite{Comlex}. Example categories include
intransitive verbs and verbs that take 2 prepositional phrases as a
complement (e.g., {\em fly} in {\em ``I fly from here to there.''}). In
addition, Comlex gives our system the possible prepositions (e.g. 
{\em from} and {\em to} for the verb {\em fly}) and particles used in the
possible subcategorizations.

The original system chose between two possible attachment points, a
verb and a noun. Each rule either attempted to move left (attach to
the verb) or move right (attach to the noun). Our extensions include
as possible attachment points every group that precedes the attaching
I-group and is in the I-group's sentence. The rules now can move the
attachment either left or right from the current guess to the nearest
group that matches the rule's constraints.

In addition to running the training and test with {\bf ALL} possible
attachment points (every preceding group) available, one can also
restrict the possible attachment points to only the group Adjacent to
the I-group and the nearest Verb group on the left, if any ({\bf
V-A}). One uses the same attachment choice ({\bf ALL} versus {\bf
V-A}) in the training run and corresponding test run.

\section{Experiments} \label{s:experiments}

\subsection{Data preparation}\label{ss:data-prep}
Our experiments were conducted with data made available through the
Penn Treebank annotation effort \cite{PennTreebank}.  However, since
our grammar model is based on syntax groups, not conventional
categories, we needed to extend the Treebank annotations to include
the constructs of interest to us.

This was accomplished in several steps.  First, noun groups and verb
groups were manually annotated using Treebank data that had been
stripped of all phrase structure markup.\footnote{We used files
200-299, along with files 2000 and 2013.}  This syntax group markup
was then reconciled with the Treebank annotations by a semi-automatic
procedure.  Usually, the procedure just needs to overlay the
syntax group markup on top of the Treebank annotations.  However, the
Treebank annotations often had to be adjusted to make
them consistent with the syntax groups (e.g., verbal auxiliaries
need to be included in the relevant verb phrase).  Some 4-5\%
of all Treebank sentences could not be automatically reconciled in
this way, and were removed from the data sets for these experiments.

The reconciliation procedure also automatically tags the data for
part-of-speech, using a high-performance tagger based on
\cite{BrillPhD}.  Finally, the reconciler introduces adjective,
adverb, and I-group markup.  I-groups are created for all lexemes
tagged with the IN, TO, WDT, WP, WP\$ or WRB parts of speech, as well
as multi-word prepositions such as {\em according to}.

The reconciled data are then compiled into attachment problems using
another semi-automatic pattern-matching procedure. 8\% of the cases
did not fit into the patterns and required manual intervention. 

We split our data into a training set (files 2000, 2013, and 200-269)
and a test set (files 270-299).  Because manual intervention is time
consuming, it was only performed on the test set. The training set
(called 0x6x) has 2615 attachment problems and the test set (called
7x9x) has 2252 attachment problems.

\subsection{Preliminary test} \label{ss:prelim-expr}
The preliminary experiment with our system compares it to previous
work \cite{Ratna94,BandR94,CandB95} when handling VNPN binary PP
attachment ambiguity. In our terms, the task is to determine the
attachment of certain {\bf vnpn} category I-groups. The data
originally was used in \cite{Ratna94} and was derived from the Penn
Treebank Wall St. Journal. It consists of about 21,000 training
examples (call this $lt$, short for {\em large-training}) and about
3000 test examples. The format of this data is slightly different than
for 0x6x and 7x9x: for each sample, only the 4 mentioned groups (VNPN)
are provided, and for each group, this data just provides the
head-word. As a result, our part-of-speech tagger could not run on
this data, so we temporarily adjusted our system to only consider two
part-of-speech categories: {\em numbers} for words with just commas,
periods and digits, and {\em non-numbers} for all other words. The
training used a 3 improvement threshold.  With these rules, the
percent correct on the test set went from 59.0\% (guess all adjacent
attachments) to 83.1\%, an error reduction of 58.9\%. This result is
just a little behind the current best result of 84.5\% \cite{CandB95}
(using a binomial distribution test, the difference is statistically
significant at the 2\% level). \cite{CandB95} also reports a result of
81.9\% for a word only version of the system \cite{BandR94} that we
extend (difference with our result is statistically significant at the
4\% level). So our system is competitive on a known task.

\subsection{The main experiments}\label{ss:main-expr}
We made 4 training and test run pairs:
\begin{center}
\begin{tabular}{|c|c||c|r|r|}\hline
TR SET & AP  & RULES & COR & ER \\ \hline \hline
0x6x & {\bf ALL}  & 159 & 70.1\% & 7.4\% \\ \hline
0x6x & {\bf V-A}  & 118 & 73.0\% & 16.3\% \\ \hline
$lt^-$ & {\bf V-A}  & 444 & 75.4\% & 24.0\% \\ \hline
$lt2^-$ & {\bf V-A}  & 451 & 74.7\% & 21.8\% \\ \hline
\end{tabular}
\end{center}
The test set was always 7x9x, which starts at 67.7\% correct. The
results report the number of RULES the training run produces, as well
as the percent CORrect and Error Reduction in the test. One source of
variation is whether {\bf ALL} or the {\bf V-A} Attachment Points are
used. The other source is the TRaining SET used.

The set $lt^-$ is the set $lt$ (Section~\ref{ss:prelim-expr}) with the
entries from Penn Treebank Wall St. Journal files 270 to 299 (the
files used to form the test set) removed. About 600 entries were
removed. Several adjustments were made when using $lt^-$: The
part-of-speech treatment in Section~\ref{ss:prelim-expr} was
used. Because $lt^-$ only gives two possible attachment points (the
adjacent noun and the nearest verb), only {\bf V-A} attachment points
were used. Finally, because $lt^-$ is much slower to train on than
0x6x, training used a 3 improvement threshold. For 0x6x, a 2
improvement threshold was used.

Set $lt2$ is the data used in \cite{Merlo97} and has about 26000
entries.  The set $lt2^-$ is the set $lt2$ with the entries from Penn
Treebank files 270-299 removed. Again, about 600 entries were
removed. Generally, $lt2$ has no information on the word(s) to the
right of the preposition being attached, so this field was ignored in
both training and test.  In addition, for similar reasons as given for
$lt^-$, the adjustments made when using $lt^-$ were also made when
using $lt2^-$.

If one removes the $lt2^-$ results, then all the COR results are
statistically significantly different from the starting 67.7\% score
and from each other at a 1\% level or better. In addition, the $lt2^-$
and $lt^-$ results are {\em not} statistically significantly different
(even at the 20\% level).

$lt2^-$ has more data points and more categories of data than $lt^-$,
but the $lt^-$ run has the best overall score. Besides pure chance,
two other possible reasons for this somewhat surprising result are
that the $lt2^-$ entries have no information on the word(s) to the
right of the preposition being attached ($lt^-$ does) and both
datasets contain entries not in the other dataset.

When looking at the $lt^-$ run's remaining errors, 43\% of
the errors were in category {\bf \={v}npn}, 21\% in {\bf vnpn}, 16\%
in {\bf x\={n}px}, 13\% in {\bf xxsx}, 4\% in {\bf \={v}np\={n}} and
3\% in {\bf vnp\={n}}.


\subsection{Afterwards}
The $lt^-$ run has the best overall score.  However, the $lt^-$ run
does not always produce the best score for each category. Below are
the scores (number correct) for each run that has a best score (bold
face) for some category:
\begin{center}
\begin{tabular}{|c||r|r|r|}\hline
Category & 0x6x ({\bf V-A}) & $lt^-$ & $lt2^-$ \\ \hline \hline
{\bf vnpn} & 345 & {\bf 397} & 374 \\ \hline
{\bf vnp\={n}} & 35 & {\bf 39} & 34 \\ \hline
{\bf \={v}npn} & 441 & 454 & {\bf 458} \\ \hline
{\bf \={v}np\={n}} & 32 & 29 & {\bf 36} \\ \hline
{\bf x\={n}px} & 554 & 551 & {\bf 557} \\ \hline
{\bf xxsx} & {\bf 236} & 229 & 224 \\ \hline
\end{tabular}
\end{center}
The location of most of the best subscores is not surprising. Of the
training sets, $lt^-$ has the most {\bf vnpn} entries,\footnote{For
{\bf vnpn}, the $lt^-$ score is statistically significantly better
than the $lt2^-$ score at the 2\% level.} $lt2^-$ has the most {\bf
\={v}np}-type entries and 0x6x has the most {\bf xxsx} entries. The
best {\bf vnp\={n}} and {\bf x\={n}px} subscore locations are
somewhat surprising. The best {\bf vnp\={n}} subscore is statistically
significantly better than the $lt2^-$ {\bf vnp\={n}} subscore at the
5\% level.  A possible explanation is that the {\bf vnp\={n}} and {\bf
vnpn} categories are closely related. The best {\bf x\={n}px} subscore
is not statistically significantly better than the $lt^-$ {\bf
x\={n}px} subscore, even at the 25\% level. Besides pure chance, a
possible explanation is that the {\bf x\={n}px} category is related to
the four {\bf np}-type categories (where $lt2^-$ has the most
entries).

The fact that the subscores for the various categories differ
according to training regimen suggests a system architecture that
would exploit this.  In particular, we might apply different rule sets
for each attachment category, with each rule set trained in the
optimal configuration for that category.  We would thus expect the
overall accuracy of the attachment procedure to improve overall.  To
estimate the magnitude of this improvement, we calculated a post-hoc
composite score on our test set by combining the best subscore for
each of the 6 categories. When viewed as trying to improve upon the
$lt^-$ subscores, the new {\bf \={v}np\={n}} subscore is statistically
significantly better (4\% level) and the new {\bf xxsx} subscore is
mildly statistically significantly better (20\% level).  The new {\bf
\={v}npn} and {\bf x\={n}px} subscores are not statistically
significantly better, even at the 25\% level.
This
combination yields a post-hoc improved score of 76.5\%.  This is of
course only a post-hoc estimate, and we would need to run a new
independent test to verify the actual validity of this effect. Also,
this estimate is only mildly statistically significantly better (13\%
level) than the existing 75.4\% score.

\section{Discussion}
This paper presents a system for attaching prepositions and
subordinate conjunctions that just relies on easy-to-find constructs
like noun groups to determine when it is applicable.  In sample text,
we find that the system is applicable for trying to attach 89\% of the
prepositions/subordinate conjunctions that are outside of the
easy-to-find constructs and is 75.4\% correct on the attachments that it
tries to handle. In this sample, we also notice that these attachments
very much tend to be to only one or two different spots and that the
attachment problems can be divided into 6 categories. One just needs
those easy-to-find constructs to determine the category of an
attachment problem.

The 75.4\% results may seen low compared to parsing results like the
88\% precision and recall in \cite{Collins97}, but those parsing
results include many easier-to-parse constructs. \cite{MandC97}
presents the VNPN example phrase {\em ``saw the man with a telescope''}, where
attaching the preposition incorrectly can still result in 80\% (4 of
5) recall, 100\% precision and no crossing brackets. Of the 4 recalled
constructs, 3 are easy-to-parse: 2 correspond to noun groups and 1 is
the parse top level.

In our experiments, we found that limiting the choice of possible
attachment points to the two most likely ones improved
performance. This limiting also lets us use the large training sets
$lt^-$ and $lt2^-$. In addition, we found that different training data
produces rules that work better in different categories. This latter
result suggests trying a system architecture where each
attachment category is handled by the rule set most suited for
that category.

In the best overall result, nearly half of the remaining errors occur
in one category, {\bf \={v}npn}, so this is the category in need of
most work.

Another topic to examine is how many of the remaining attachment
errors actually matter. For instance, when one's interest is on
finding a semantic interpretation of the sentence {\em ``They flash
letters on a screen.''}, whether {\em on} attaches to {\em flash} or
to {\em letters\/} is irrelevant. Both the {\em letters} are, and the
{\em flash\/}ing occurs, {\em on a screen}.

\end{document}